\pdfoutput=1

\documentclass[11pt]{article}

\usepackage[final]{acl}

\usepackage{times}
\usepackage{latexsym}

\usepackage[T1]{fontenc}

\usepackage[utf8]{inputenc}

\usepackage{microtype}

\usepackage{inconsolata}

\usepackage{graphicx}

\usepackage{booktabs}
\usepackage{xtab}
\usepackage{algorithm, algorithmic}
\usepackage{amsfonts}
\usepackage{amsmath}
\usepackage{tikz}
\usepackage{tabularx}
\usepackage{makecell}
\usepackage{todonotes}
\usepackage{float}
\newcolumntype{L}[1]{>{\raggedright\arraybackslash}p{#1}}
\newcommand*\emptycirc[1][1ex]{\tikz\draw (0,0) circle (#1);} 
\newcommand*\halfcirc[1][1ex]{%
  \begin{tikzpicture}
  \draw[fill] (0,0)-- (90:#1) arc (90:270:#1) -- cycle ;
  \draw (0,0) circle (#1);
  \end{tikzpicture}}
\newcommand*\fullcirc[1][1ex]{\tikz\fill (0,0) circle (#1);}

\usepackage{xcolor}

\usepackage{url}

\usepackage{tcolorbox}
\tcbuselibrary{skins}

\usepackage{tikz}
\usetikzlibrary{positioning, chains}

\usepackage{soul}

\title{Ensemble Watermarks for Large Language Models}

\author{First Author \\
  Affiliation / Address line 1 \\
  Affiliation / Address line 2 \\
  Affiliation / Address line 3 \\
  \texttt{email@domain} \\\And
  Second Author \\
  Affiliation / Address line 1 \\
  Affiliation / Address line 2 \\
  Affiliation / Address line 3 \\
  \texttt{email@domain} \\}

\author{
  \textbf{Georg Niess\textsuperscript{1}},
  \textbf{Roman Kern\textsuperscript{1,2}},
\\
\\
  \textsuperscript{1}Graz University of Technology,
  \textsuperscript{2}Know Center Research GmbH,
\\
  \small{
    \textbf{Correspondence:} \href{mailto:georg.niess@tugraz.at}{georg.niess@tugraz.at}
  }
}

\begin{document}
\maketitle
\begin{abstract}
As large language models (LLMs) reach human-like fluency, reliably distinguishing AI-generated text from human authorship becomes increasingly difficult.
While watermarks already exist for LLMs, they often lack flexibility and struggle with attacks such as paraphrasing.
To address these issues, we propose a multi-feature method for generating watermarks that combines multiple distinct watermark features into an ensemble watermark.
Concretely, we combine acrostica and sensorimotor norms with the established red-green watermark to achieve a 98\% detection rate.
After a paraphrasing attack, the performance remains high with 95\% detection rate. 
In comparison, the red-green feature alone as a baseline achieves a detection rate of 49\% after paraphrasing.
The evaluation of all feature combinations reveals that the ensemble of all three consistently has the highest detection rate across several LLMs and watermark strength settings. 
Due to the flexibility of combining features in the ensemble, various requirements and trade-offs can be addressed. 
Additionally, the same detection function can be used without adaptations for all ensemble configurations.
This method is particularly of interest
to facilitate accountability and prevent societal harm.
\end{abstract}

\section{Introduction}

The inception of transformer-based architectures \citep{vaswani_attention_2023} combined with large-scale pertaining \citep{devlin_no_2019,radford_language_2019} has continuously improved the performance of modern large language models (LLMs). Humans increasingly struggle to distinguish between texts written by LLMs and those created by humans, as machine-generated texts sometimes even deceive people more often than human-written ones \citep{zellers_defending_2019}. Although a growing range of post-hoc detectors exists, many detection methods that were once considered reliable for GPT-2 now struggle with GPT-3 and later versions~\citep{fagni_tweepfake_2021}.
This arms race between generation and detection intensifies as model size and capabilities continue to scale \citep{kaplan_scaling_2020}.
Several potential cases of misuse are already associated with these advanced models \citep{ray_chatgpt_2023}.

Watermarks attempt to solve this by embedding a secret code into LLM output by modifying logits of the generated tokens during the generation process. However, as we will show later, a watermark with only a single feature has limited resilience against attacks like paraphrasing. 
To help against this weakness, we introduce an ensemble watermark that combines stylometric watermark features like acrostica and sensorimotor norms with the established red-green watermark feature introduced by~\citet{kirchenbauer_watermark_2023}.
Our method is flexible and allows for a diverse set of features, and we draw inspiration from features used in the context of stylometry. 

Stylometry is the statistical analysis of variations in literary style, with numerous stylometric features proposed in literature \cite{neal_surveying_2018}.
These features serve as a writer's fingerprint and include aspects such as syntax, vocabulary, sentence structure, sentence length, and other unique authorial characteristics. 
All of these features can be statistically analyzed and utilized in tasks like authorship attribution~\citep{stamatatos_survey_2009}.
However, traditional authorship attribution techniques have shown challenges when applied to even smaller language models like GPT-2~\cite{uchendu_authorship_2020}. 

Sensorimotor norms are classifications based on human cognition. 
Perceptual modalities, such as "hearing", and action effectors, such as "hand", have been extensively researched in psychology and cognitive semantics ~\cite{lynott_lancaster_2020}. However, there is still limited research in computer science on this topic.
In our approach, a secret key derived from the generated output selects the sensorimotor category, thereby influencing the generated words. For example, "smells funny" would be preferred over "looks funny" for the olfactory category.

An acrostic is a text in which the first letter of each sentence can be combined to spell out a hidden message or word. 
Historically, authors have used acrostics to encode their authorship~\cite{johnson2006authorial}, often incorporating variations of their names into the hidden message. A notable example of an acrostic can be found in Appendix~\ref{tab:acrostic-schwarz}.
In our method, the secret key determines which letters are used as the first letter of the first word in each generated sentence.

The main contributions of this paper are as follows:
\begin{itemize}
    \itemsep0em
    \item We propose an ensemble watermark approach for LLMs based on changing token logits on a token and sentence-based level to embed novel stylometric features combined with an established red-green watermark.
    \item We show that this provides more resilience against paraphrasing attacks for three LLMs and three different parameter settings and provides the best detection rate.
    \item We propose a detection method that works for any combination of our ensemble watermark features, even in isolation, without any changes to the function.
\end{itemize}

The flexible nature of our ensemble watermark allows it to be adapted to different requirements while using the same detection method. All code and data generated is available on GitHub\footnote{\url{https://github.com/CommodoreEU/ensemble-watermark}}.

\section{Background \& Related Work}

\begin{table*}[htbp] 
\small
\caption{
Overview of common stylometric features as identified in literature.
Their suitability for four distinct ways to integrate watermarks into LLMs are reported based on a self-assessment. 
A black circle shows the best compatibility, followed by the semi-filled circle. 
For our ensemble we decided on sensorimotoric words and acrostics next to the established red-green feature, and on logit manipulation for watermarking. (Extended from \citealp{niess2024stylometricwatermarkslargelanguage})
 }
    \vspace{-0.2cm}
\begin{minipage}[b]{0.45\linewidth}
    \begin{tabular}{lcccc}
    \toprule
    \textbf{Feature Type} & \textbf{Logits} & \textbf{Fine-Tuning}  & \textbf{Prompt} & \textbf{Post-Processing}  \\
    \midrule
    N-Grams                 & \fullcirc & \fullcirc & \emptycirc & \emptycirc \\
    Character Frequency     & \fullcirc & \halfcirc & \halfcirc & \halfcirc \\
    Vocabulary Richness     & \fullcirc & \fullcirc & \halfcirc & \fullcirc \\
    Word Distributions      & \halfcirc & \fullcirc & \halfcirc & \halfcirc \\
    Word Length             & \fullcirc & \fullcirc & \halfcirc & \halfcirc \\
    Sentence Length         & \halfcirc & \fullcirc & \emptycirc & \emptycirc \\
    Parts of Speech         & \emptycirc& \fullcirc  & \emptycirc & \emptycirc  \\
    Punctuation Frequency   & \halfcirc & \halfcirc & \emptycirc & \halfcirc  \\
    Sentence Complexity     & \halfcirc & \fullcirc  & \halfcirc & \halfcirc   \\
    Synonyms                & \emptycirc& \halfcirc  & \emptycirc & \fullcirc  \\
    \midrule
    Sensorimotoric Words    & \fullcirc & \fullcirc & \fullcirc & \emptycirc  \\
    Acrostics               & \fullcirc & \emptycirc& \halfcirc & \fullcirc  \\
    Red-Green               & \fullcirc & \emptycirc& \emptycirc & \halfcirc  \\
    \bottomrule
    \end{tabular}\\
\end{minipage}
\hspace{1cm}
\begin{minipage}[b]{0.45\textwidth}
    \hspace{3cm}%
    \vspace{-0.2cm}
    \includegraphics[width=4.5cm]{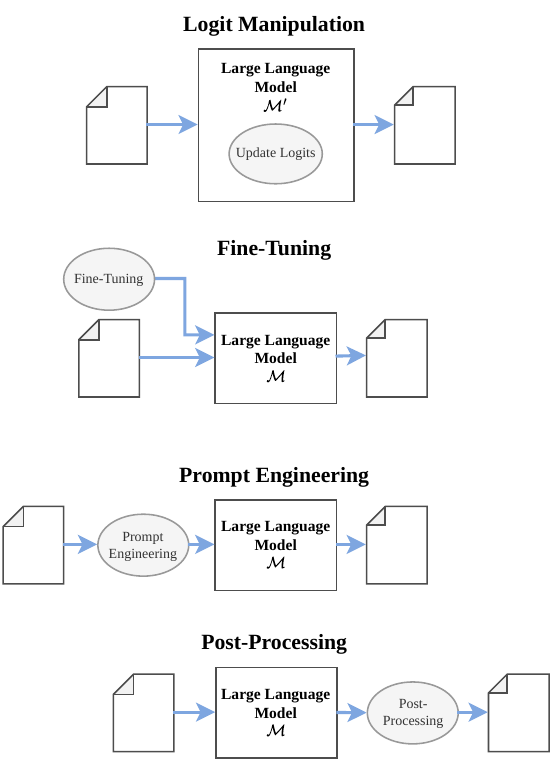}\\
\end{minipage}
\label{table:stylometricFeaturesCompatibility}
\end{table*}

Several notable attempts have been made to detect machine-generated text without using language and stylometric features. Depending on their functional approach, these efforts can generally be categorized into two main groups.

\subsection{Watermarking Approaches}

\paragraph{Post-hoc detection.}The techniques in this category aim to identify machine-generated text without adding watermarks or altering either the LLM itself or its output. Notable examples of methods in this include the classifier developed by OpenAI~\cite{kirchner_new_2023}, GPTZero~\cite{tian_gptzero_2023}, and DetectGPT~\citep{mitchell_detectgpt_2023} which utilizes the probability curvature of text sampled from an LLM. DetectGPT relies on the principle that text typically exists within negative curvature regions of a model's log probability. This means that even minor changes to a sentence will decrease its log likelihood, as an LLM continuously aims to optimize the probability of each sentence. Additional post-hoc methods can be found in the survey by~\citet{jawahar_automatic_2020}. One advantage of post-hoc detection methods is that they can be applied to any suspected text without prior requirements. However, they are vulnerable to user attacks and become less effective with more complex language models \citep{chakraborty_possibilities_2023}.

\paragraph{Watermarking and Red-Green Watermarks.}
Watermarking is the technique of hiding information in data that is difficult for others to remove but can be detected by an algorithm to read the hidden information. 
Some successful watermark implementations already exist, such as \cite{kirchenbauer_watermark_2023}, \cite{christ_undetectable_2023}, and there is ongoing research by \cite{aaronson_my_2022, aaronson_watermarking_2023}.
The work by \cite{kirchenbauer_watermark_2023}, which we also include in our ensemble watermark and all our comparisons, divides tokens into two categories: a green list and a red list. Tokens in the green list are given a weight boost to enhance their representation in the generated output. As a result, the generated text predominantly consists of words from the green list, while human-written text naturally incorporates words from the red list.
Additionally, \citet{xiang2024reversible} tackle the challenge of replacing sensitive words by utilizing Sentence-BERT~\citep{reimers2019sentence}.
\citet{zhao_provable_2024} also use red-green watermarking, but instead of dynamically generating the red-green list, use a fixed list to simplify the grouping strategy. \citet{kuditipudi_robust_2024} then test several sampling schemes to improve watermark performance. 
The Duwak approach~\cite {zhu_duwak_2024} combines the red-green watermark feature with a sampling scheme to improve watermarking; it does, however, not combine multiple logit watermark features as we do.

\subsection{Stylometric Features}

A wide variety of stylometric features have been proposed in the literature \citep{lagutina_survey_2019,stamatatos_survey_2009}. 
These features can be broadly classified into five categories: lexical, syntactic, semantic, structural, and domain-specific features. 
In Table~\ref{table:stylometricFeaturesCompatibility}, we have compiled a list of popular stylometric features found in the literature, along with additional watermark features we consider promising.

\paragraph{Sensorimotor Norms.}
\citet{winter2019sensory} define sensory linguistics as the study of the relationship between senses and language.
The Lancaster Sensorimotor Norms~\citep{lynott_lancaster_2020} detail a set of 40,000 sensorimotor words, accompanied by a crowd-based assessment across 11 dimensions, which serve as the foundation for our sensorimotor features. 
Every word can be categorized into six perceptual groups: touch, hearing, smell, taste, vision, and interoception, as well as five action categories: mouth/throat, hand/arm, foot/leg, head (excluding mouth/throat), and torso. \citet{khalid2022smells} utilize the Lancaster Norms and conclude that sensorial language is likely used intentionally and should not be viewed as a random phenomenon. Additionally, perceptual features have been suggested in the literature for authorship attribution, particularly in cross-language contexts ~\citep{bogdanova_cross-language_2014}.

\paragraph{Acrostica.}
\citet{stein2014generating} explored the task of generating paraphrased versions of existing texts that also include acrostics.
They approached this challenge as a search problem.
\citet{shen2019controlling} employed a sequence-to-sequence network to generate texts with acrostics in both English and Chinese. More recently, using steganography for embedding secret messages in text has been investigated, utilizing BERT~\citep{yi2022alisa}.

\subsection{Watermarking Implementations}

LLMs can utilize watermarks, including stylometric ones, in several ways. We present four different approaches in the following. 
They have different strengths and weaknesses. We outline four distinct approaches below, each with its own strengths and weaknesses. Additionally, we provide an overview of these methods and assess their expected effectiveness for different stylometric features in Table~\ref{table:stylometricFeaturesCompatibility}.

\paragraph{Logits Manipulation.}
 In our approach, we have opted for directly manipulating the logits of the LLM tokens generated. Logits are the raw output values of a machine learning model before applying an activation function like softmax. They represent the unnormalized score or prediction for each token.
 This manipulation gives considerable control over how often single features are generated, though the difficulty lies in finding the correct logits that produce the desired features. 
 An example of this type of watermark can be found in the work by \citet{kirchenbauer_watermark_2023}.

\paragraph{Fine-tuning.}
A standard solution is to conduct additional training to change the output of an LLM. This method can also be applied to watermarks, although it is less controlled than other techniques. For instance, the written works of an author with a distinctive writing style can be used to fine-tune an LLM to produce text resembling that author's style~\cite{li_teach_2023}. 
However, a significant challenge is controlling what the model learns. The LLM might not only pick up the desired watermark features but could also overfit to unrelated aspects, leading to a loss of generalization, domain mismatch, and decreased robustness in generating text.

\paragraph{Prompt Engineering.}
Another alternative is to use the inherent capabilities of LLMs by designing a system prompt that specifies the desired features~\cite{zhou_large_2023}. For example, it is possible to tell contemporary LLMs to use only certain letters in their sentences~\cite{openai_chat_2024b}. The primary challenges associated with this method include creating an effective prompt, protecting against users writing their own prompt, and understanding what prompts the language model can comprehend and adhere to. Generally, more powerful models are better suited for this approach.

\paragraph{Post-Processing.}
The final option is to process the text after generation is complete. This is how early attempts were made by \citet{topkara_natural_2005} and \citet{goos_natural_2001} before LLMs were developed. It was commonly used to embed watermarks for copyright and document integrity purposes. Although the generative capabilities of LLMs have diminished the appeal of this approach, it can still be beneficial for implementing simple features like synonym replacement, which do not require generating new sentences.

\section{Method}

Our approach includes two main components: 1) a process for generating watermarks by manipulating logits with dynamic keys, and 2) a test procedure for detecting an existing watermark using statistical tests.

\subsection{Watermark Generation} \label{construction}

The generation algorithm modifies the logits of the language model during text generation to embed the features forming the watermark. 
It adjusts token probabilities to make certain words more likely, based on: 
The acrostic pattern, by boosting tokens that start with specific letters. 
Sensorimotor words, by boosting tokens associated with a target sensorimotor class. 
The red-green mechanism, by adjusting token probabilities based on a dynamically generated green list.
We follow the procedure proposed in the original paper~\cite{kirchenbauer_watermark_2023} for splitting the red-green list.
For the other features, a secret key is required, as described below.

\begin{algorithm}[h]
\caption{Watermarked Text Generation}\label{alg:generate}
\begin{algorithmic}[1]
\STATE Initialize secret key: $\text{senso\_class}, \text{acro\_letter}$
\STATE Set $\delta_{\text{acro}}, \delta_{\text{senso}}, \delta_{\text{redgreen}}$
\WHILE{not done generating}
    \STATE Get current logits from the model
    \IF{starting a new sentence}
        \STATE Adjust logits for acrostic boosting
    \ELSE
        \STATE Adjust logits for sensorimotor boosting
    \ENDIF
    \STATE Generate green list based on last token
    \STATE Adjust logits for red-green mechanism
    \STATE Sample next token from adjusted logits
    \STATE Update the secret key based on last word or sentence
\ENDWHILE
\end{algorithmic}
\end{algorithm}
\paragraph{Secret Key Generation.}

A secret key is maintained throughout the generation process to control most of the features, i.e., it determines the sensorimotor class and the letter for the acrostic pattern. 
The key is updated based on the last word and the last sentence using secure hash functions. Both words and sentences are hashed using the same base function. 
Given a word $w$, the hash function maps it to an integer directly within a specified range $[a, b]$. 
Given a sentence $s$, we first lemmatize and remove stopwords and punctuation to get a sentence $s'$. 
The hash function is applied to $s'$ to generate an integer within a range $[a', b']$.

\begin{align*}
\text{hash}(x) &= \left( \text{int}\left( \text{SHA256}(x) \mod 2^{32} \right) \right. \\
& \left. \mod (b - a + 1) \right) + a
\end{align*}

\paragraph{Logits Adjustment.} During generation, the logits (raw scores before softmax) are adjusted to boost the probability of specific tokens. 
For example, if a new sentence is started, the initial token is boosted according to the target acrostic letter. 
For each token $t$:
\[
\text{logits}[t] += \delta_{\text{acro}} \cdot \mathbf{1}\{\text{starts\_with\_acrostic\_letter}\}
\]
Otherwise, boost tokens that are associated with the current sensorimotor class. For each token $t$:
\[
\text{logits}[t] += \delta_{\text{senso}} \cdot \mathbf{1}\{\text{token\_in\_sensorimotor\_class}\}
\]

The red-green mechanism is based on the work by~\cite{kirchenbauer_watermark_2023}. A green list is generated based on the last token $t_{\text{last}}$. A random number is then seeded by a generator with: $\text{seed} = \text{hash}(t_{\text{last}})$. This is used to generate a random permutation of the vocabulary, where the first $\gamma \cdot V$ tokens are selected as the green list, where $V$ is the vocabulary size and $\gamma$ is a predefined proportion (e.g., $0.5$) For tokens in the green list: $\text{logits}[t] += \delta_{\text{redgreen}}$

It is important to ensure consistent tokenization between generation and detection and appropriately handle special tokens (e.g., BOS, EOS). Streaming generation is used to update the secret key and adjust logits at each step during generation. To detect sentence boundaries, punctuations are used (e.g., \texttt{.}, \texttt{!}, \texttt{?}).

We decided to assign a relatively large weight boost to acrostica because the beginning of the sentence is more flexible, allowing the rest of the sentence to adapt to this change easily.
In contrast, we chose a small weight for sensorimotor words to prevent the model from becoming overly biased.

All features have in common that their strength is controlled via a $\delta$ parameter. 
In the evaluation we study the impact of these parameters on the generated text.
While we opted for fixed values for the weights, they could also be chosen dynamically, depending on how long it has been since a desired feature was chosen or on the distribution of the current weights.

\subsection{Watermark Detection} \label{detection}
The detection of the watermark works similarly to that of the generation. 
A secret key is maintained the same way, but instead of modifying logits, the generated token is compared based on the key, and the probability of that token occurring is calculated. 

\begin{algorithm}
\caption{Watermark Detection Algorithm}\label{alg:detection}
\begin{algorithmic}[1]
\STATE \textbf{Input}: Text $T$
\STATE \textbf{Initialize}: Load sensorimotor norms and class frequencies from the corpus
\STATE Initialize secret key: \\$\text{sensorimotor\_class}, \text{acrostic\_letter}$
\STATE Initialize counters for acrostic matches $k$, total checks $n$
\STATE Initialize counters for sensorimotor matches per class $k_c$, total words per class $n_c$
\STATE Initialize variables for red-green detection: total transitions $T$, green tokens $G$
\STATE Split text $T$ into sentences $S = [s_1, s_2, \dots, s_N]$
\FOR{each sentence $s_i$ in $S$}
    \STATE Split $s_i$ into words $W = [w_1, w_2, \dots, w_m]$
    \IF{$i > 1$}
        \STATE \textbf{Acrostic Check}: Compute expected letter using $\text{hash\_sentence}(s_{i-1})$
        \STATE Compare with first letter of $w_1$
        \STATE Increment $k$ if match found
    \ENDIF
    \FOR{each word $w_j$ in $W$}
        \STATE Update sensorimotor class using $\text{hash\_word}(w_{j-1})$
        \STATE $c \leftarrow \text{sensorimotor\_class}$
        \IF{$w_j$ is in sensorimotor dictionary}
            \STATE $n_c \leftarrow n_c + 1$
            \IF{$w_j$ belongs to class $c$}
                \STATE $k_c \leftarrow k_c + 1$
            \ENDIF
        \ENDIF
        \STATE \textbf{Red-Green Detection}: Update green list based on previous token
        \STATE Increment $G$ if $w_j$ is in green list
    \ENDFOR
\ENDFOR
\STATE \textbf{Calculate Probabilities}: Compute acrostic probability $P_{\text{acrostic}}$
\STATE Compute sensorimotor probability $P_{\text{sensorimotor}} = \prod_{c} P_{\text{sensorimotor}, c}$
\STATE Compute red-green probability $P_{\text{redgreen}}$
\STATE \textbf{Compute Final Score}:
\[
\text{final\_score} = P_{\text{acrostic}} \times P_{\text{sensorimotor}} \times P_{\text{redgreen}}
\]
\STATE \textbf{Output}: $\text{final\_score}$
\end{algorithmic}
\end{algorithm}

\begin{figure*}[t]
    \vskip 0.2in
    \begin{center}
    \centerline{\includegraphics[width=.95\textwidth]{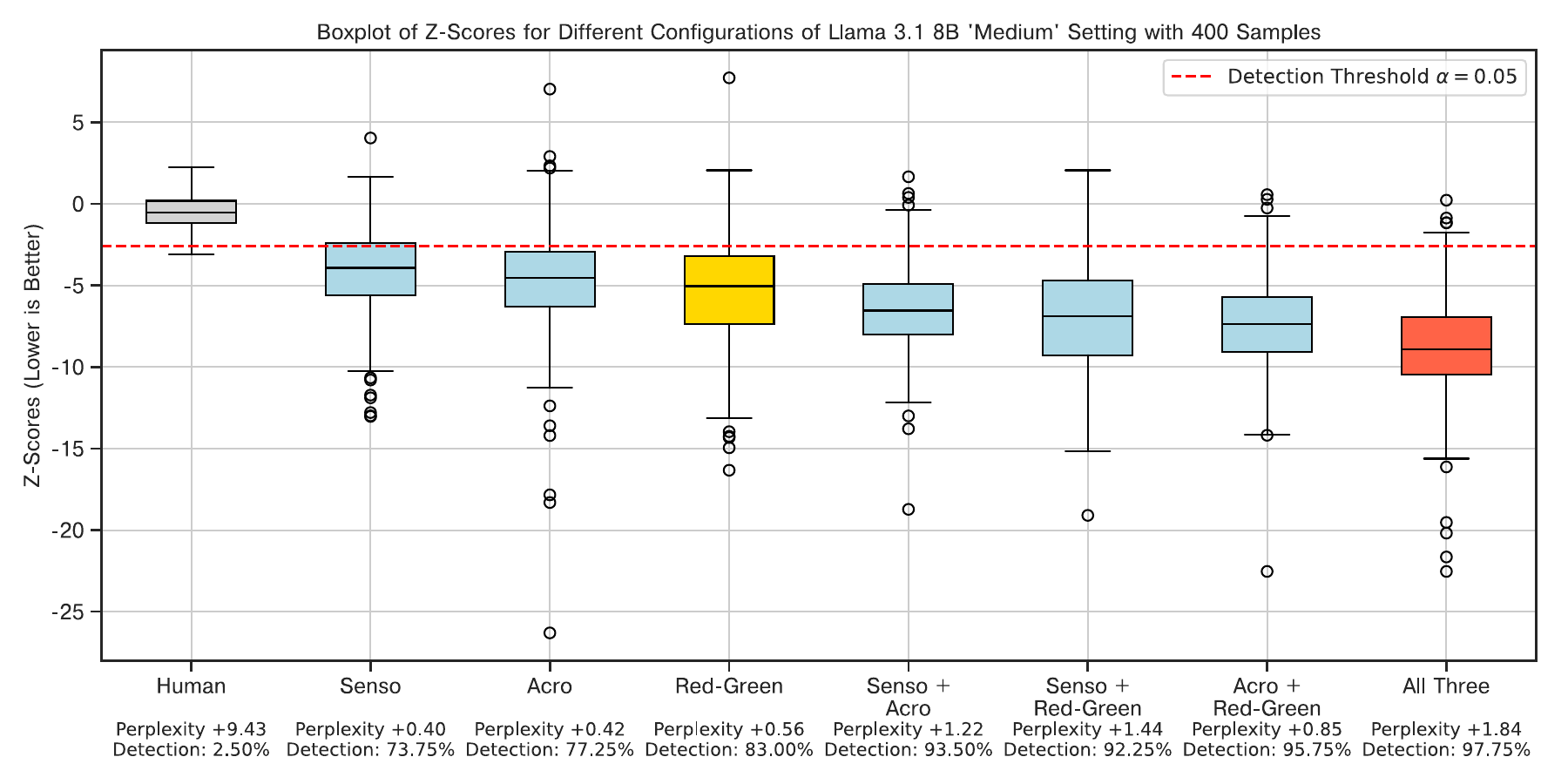}}
    \caption{Plot of the Z-Scores of different configurations of the watermark ensemble (Llama 3.1 8B,  medium strength watermark). The human and red-green results serve as baselines.
    Each watermark configuration has 400 unique samples. The combination of all features achieves the best detection rate of 97.75\%. The combination of acrostic and red-green might be attractive for its more moderate increase in perplexity (+0.85), while still achieving a detection rate of 95.75\%. The red-green baseline achieves a detection rate of 83.00\%.
    }
    \label{fig:medium}
    \end{center}
    \vskip -0.2in
\end{figure*}

The probability of detecting an acrostic watermark is calculated using the formula:
\[
P_{\text{acrostic}} = 1 - \sum_{i=0}^{k - 1} \binom{n}{i} p^{i} (1 - p)^{n - i}
\]
where \( n \) is the number of acrostic checks (total sentences minus one), \( k \) is the number of correct matches observed, and \( p = \frac{1}{26} \) represents the probability of a random match.

The probability of detecting a sensorimotor watermark uses class-specific frequencies from the Google N-gram corpus or evenly split probabilities. For each class \( c \), the baseline probability \( p_c \) is calculated. Given \( k_c \) matches out of \( n_c \) words for class \( c \), the probability is:
\[
P_{\text{sensorimotor}, c} = 1 - \sum_{i=0}^{k_c - 1} \binom{n_c}{i} p_c^{i} (1 - p_c)^{n_c - i}
\]
The combined probability across all classes is:
\[
P_{\text{sensorimotor}} = \prod_{c} P_{\text{sensorimotor}, c}
\]

The probability calculation for the red-green watermark involves the following parameters: \( T \), which represents the total number of transitions; \( G \), the number of tokens in the green list observed; and \( \gamma = 0.5 \), the probability parameter.
The expected number \( E \) and variance \( \text{Var} \) are:
\[
E = \gamma T, \quad \text{Var} = T \gamma (1 - \gamma)
\]
The Z-score and probability is calculated as:
\[
Z = \frac{G - E}{\sqrt{\text{Var}}}, \quad
P_{\text{redgreen}} = 1 - \Phi(Z)
\]
where \( \Phi \) is the cumulative distribution function of the standard normal distribution.
Assuming independence, the final score is:
\[
\text{final\_score} = P_{\text{acrostic}} \times P_{\text{sensorimotor}} \times P_{\text{redgreen}}
\]

\section{Evaluation}\label{evaulation}

In this section, we describe the experiments conducted, assess the resilience against a paraphrasing attack, and conduct an ablation study to analyze the impact of text length. 

\subsection{Implementation Details.} 
We used three LLMs, Llama 3.1 8B~\citep{dubey2024llama}, Llama 3.2 3B, and Mistral 7B~\cite{jiang_mistral_2023}. We utilized the logits processor of Hugging  Face~\cite{wolf_huggingfaces_2019} to manipulate the generation. To create prompts with human baselines, we randomly select texts from the C4 RealNewsLike dataset~\citep{2019t5}, trim a fixed length of tokens as "baseline" completions from the end, and treat the remaining tokens as the prompt.

We create a weak, middle, and strong parameter setting to test different parameters for the features. The medium setting [$\delta_{senso}=2.5, \delta_{acro}=20.0, \delta_{redgreen}=2.0$] has 400 samples per configuration, while the strong [$\delta_{senso}=5.0, \delta_{acro}=40.0, \delta_{redgreen} = 10.0$] and weak [$\delta_{senso}=1.0, \delta_{acro}=10.0, \delta_{redgreen}=1.0$] setting contain 300 samples per configuration.

We use a level of $\alpha = 0.05$ to determine the statistical significance that a watermark can be recovered. This allows us to calculate a detection rate showing how many samples would have been detected as a watermark with this $\alpha$ threshold.

\begin{figure}[t] 
    \vskip 0.2in
    \begin{center}
    \centerline{\includegraphics[width=0.5\textwidth]{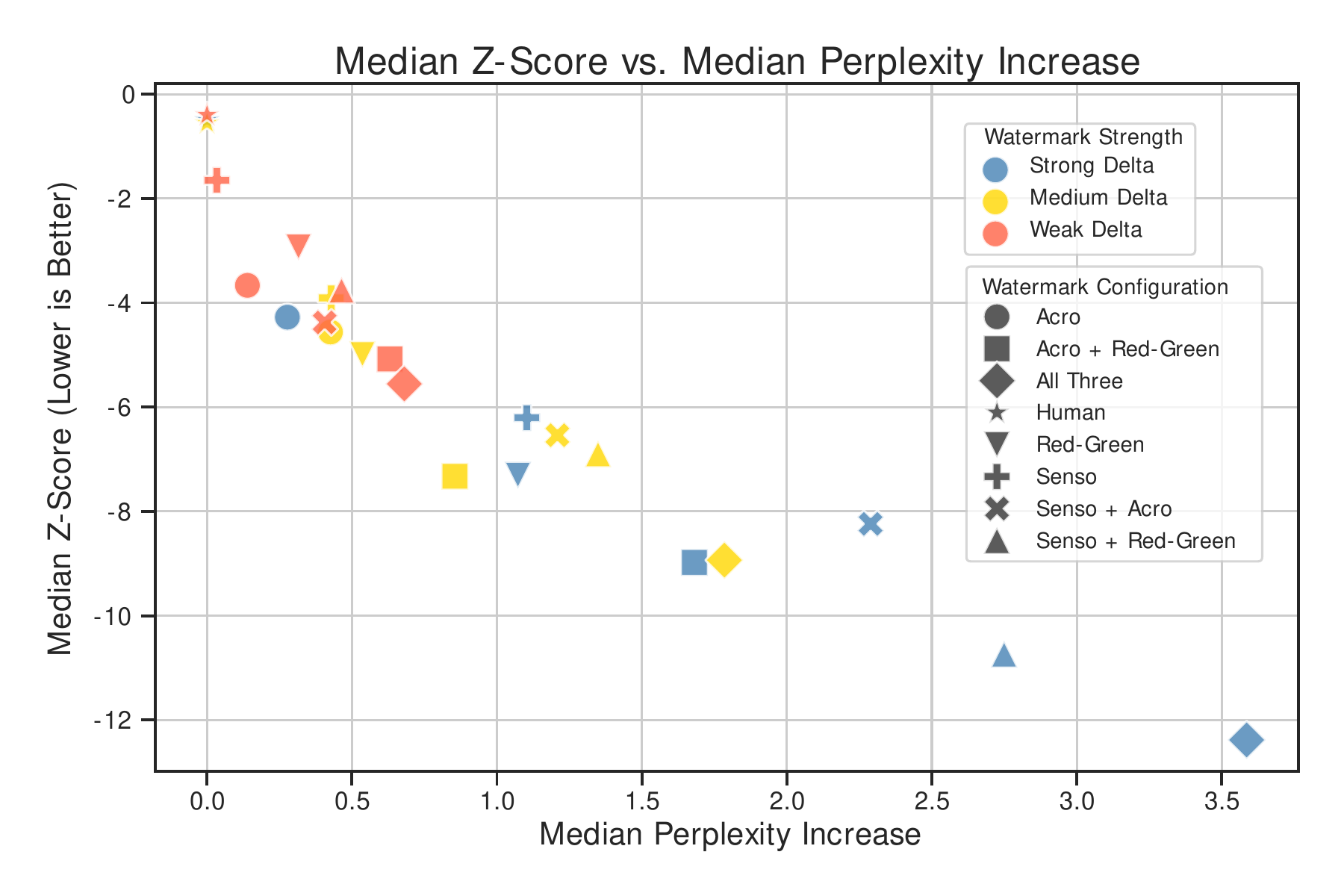}}
    \caption{Llama 3.1 8B tested with three different parameter settings. Higher parameter values for the $\delta$ weights lead to lower Z-Scores (improved detectability) and an almost linear increase in the perplexity. The acrostic feature stands out for having the least impact on perplexity, with the ensemble of all features providing the best detectability for each parameter setting.
    }
    \label{fig:perplexity}
    \end{center}
    \vskip -0.2in
\end{figure}

\subsection{Results}
Figure~\ref{fig:medium} provides an overview of combinations of features for Llama 3.1 8B, with a lower Z-Score indicating improved detectability.
Overall, combinations of watermark features have a higher detection rate than single features. 
Of the single features, the red-green feature provides the best detectability.  
The combination of all features has the best detection rate, with an increase of 17.78\% compared to the red-green feature. We also tested for statistical significance in Figure~\ref{fig:significance}. This finding is consistent across all experiments we conducted and reported in Appendix~\ref{sec:appendix} for all settings.

\begin{figure}[t]
    \vskip 0.2in
    \begin{center}
    \centerline{\includegraphics[width=0.395\textwidth]{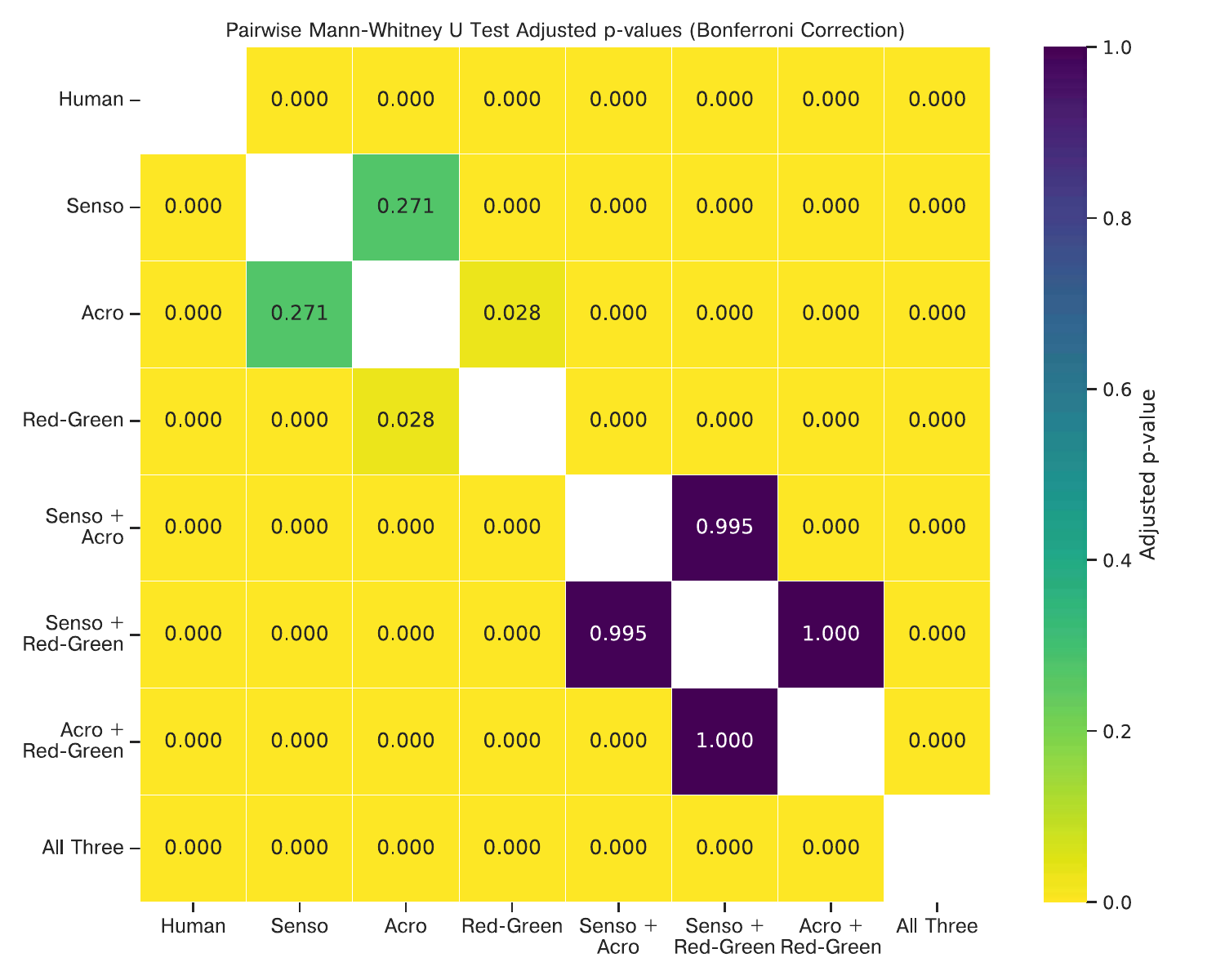}}
    \caption{Heatmap of adjusted p-values from pairwise Mann-Whitney U tests between configurations (Bonferroni correction applied) for Llama 3.1 8B with the medium parameter setting.
    Lower adjusted p-values indicate a significant difference after correction for multiple comparisons.
    }
    \label{fig:significance}
    \end{center}
    \vskip -0.2in
\end{figure}

\begin{table*}[t]
    \centering
    \small
    \begin{tabular}{lcccccccc}
        \toprule
        \makecell{Configuration} & \makecell{Human} & \makecell{Senso} & \makecell{Acro} & \makecell{Red-Green} & \makecell{Senso +\\Acro} & \makecell{Senso +\\Red-Green} & \makecell{Acro +\\Red-Green} & \makecell{All Three} \\
        \midrule
        \makecell{Llama 3.1 8B (Strong)} & 0.34 & 80.41 & 28.52 & 49.14 & 89.35 & 84.19 & 55.67 & \textbf{95.19} \\
        \makecell{Llama 3.2 3B (Strong)} & 0.97 & 85.11 & 31.39 & 54.05 & 90.61 & 90.29 & 64.08 & \textbf{95.79} \\
        \makecell{Llama 3.1 8B (Medium)} & 2.42 & 58.06 & 28.23 & 34.14 & 73.92 & 70.70 & 46.77 & \textbf{82.53} \\
        \makecell{Mistral 7B (Medium)} & 1.44 & 54.87 & 42.60 & 23.47 & 69.31 & 58.84 & 40.07 & \textbf{73.65} \\
        \makecell{Llama 3.2 3B (Weak)} & 1.32 & 26.07 & 27.72 & 29.04 & 38.61 & 43.56 & 35.97 & \textbf{44.88} \\
        \bottomrule
    \end{tabular}
    \caption{Detection rate of different LLM configurations after a paraphrasing attack changing at least 10\% of each text. All three watermark features combined are consistently the best configuration. Interestingly, Mistral 7B has better results for the acrostic feature since it tended to produce more numerous shorter sentences. 
    }
    \label{tab:attack}
\end{table*}

The scatter plot in Figure~\ref{fig:perplexity} additionally shows the three considered parameter settings (weak, medium, strong), together with the increase in perplexity.
Higher perplexity reflects more substantial deviations in the output distribution.
For all parameter settings, combining all features provides the best detectability but with an associated increase in perplexity.
Interestingly, the acrostic feature is associated with the smallest increase in perplexity, i.e., the most minor influence on the distribution of overall tokens.

\paragraph{Experiments on Paraphrasing Attacks.} 
We conducted experiments to assess the vulnerability of the various watermarking configurations via paraphrasing attacks. This experiment involves tokenizing the watermarked text, then iteratively replacing one word with a \texttt{<mask>}, and using T5~\citep{2019t5} to generate candidate replacement sequences via beam search. If one candidate differs from the original string, the attack succeeds, replacing the span with the new text. This is repeated until at least 10\% of the watermarked text is replaced.
The results can be found in Table~\ref{tab:attack} and show that the combination of all three features has the highest detection rate after the paraphrasing attack for all of the considered LLMs and parameter settings. 
Llama 3.1 3B with the strong parameter setting retains over 95\% detection rate, in contrast to the red-green feature in isolation, which drops to 49\%. 
Here, the sensorimotor feature stands out, being more resilient to this type of attack retaining over 80\% detection rate. 
The acrostic feature alone does not appear to be resilient enough to the attack and only achieves a low detection rate. However, it still contributes to overall watermark resilience, which can be seen in the improvement of the ensemble of all features against using only the combination of sensorimotor and red-green feature.
Overall, using multiple features together contributes to much greater resilience against paraphrasing attacks.

\subsection{Ablation Study Text Length}
\begin{figure*}[t]
    \vskip 0.2in
    \begin{center}
    \centerline{\includegraphics[width=.9\textwidth]{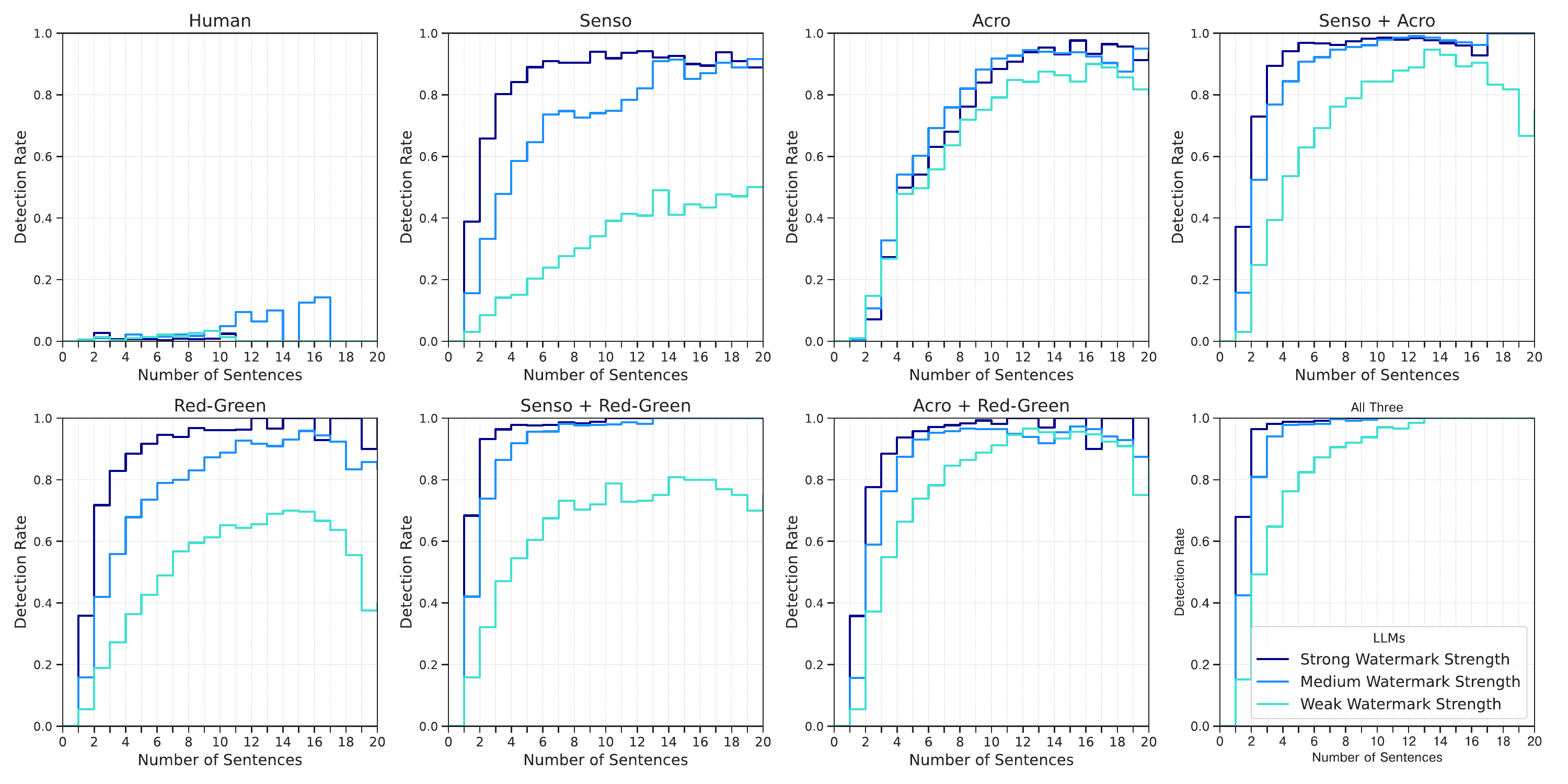}}
    \caption{For the ablation study, we study the influence of the number of output sentences on the detection rate. 
    We selected Llama 3.1 8B as language model, and present results for the weak, middle and strong parameter settings. The ensemble of all three features provides the best overall detection rate with the fewest sentences.}
    \label{fig:ablation}
    \end{center}
    \vskip -0.2in
\end{figure*}

To further analyze the influence of text length on the contribution of each feature, we do incremental sentence partitioning, where each version represents a progressively shorter subset of each text, starting with the total sentence count and reducing by one until reaching a single sentence. In Figure~\ref{fig:ablation}, the detection rate of each feature can be seen for three different watermark configurations. While acrostics are not as sensitive to differences in the $\delta$s, both sensorimotor and red-green features are associated with a jump in detection rate once an increase in the weights is applied to the logits. It can also be seen that a combination of all three features reaches a high detection rate earlier than other configurations and that the weak setting even beats some of the medium-strength configurations, hinting that for specific requirements and trade-offs specific combinations in the ensemble are best suited.

\section{Conclusion}
We introduced a novel ensemble watermarking method for large language models, specifically generative transformer models. Among the various approaches for integrating watermarks into LLMs, we chose to manipulate the probabilities directly when generating tokens.
Following existing work, we generate a key dynamically derived directly from the generated text to control the watermark features.
Out of many possible stylometric features, we focused on two features for our evaluation, the acrostic and sensorimotor norms, and combined them with an established method from literature, red-green lists. 

For the experiment, we selected three different LLMs and three levels of watermark strength.
In the evaluation, we found that each of the three features has different characteristics in relation to the detectability and perplexity of the generated sentences.

The acrostic feature has the most negligible impact on perplexity, but its performance depends on sentence number and length and has less resilience against paraphrasing than other features. The sensorimotor feature is similar in watermark performance and perplexity impact as the established red-green feature but shows increased resilience against paraphrasing. On the other hand, the red-green feature shows a balanced performance and combines well with the other (stylometric) features.

Overall, we propose a flexible and resilient ensemble watermark for text that works with short text lengths and does not require a different sampling strategy, expensive additional model training, or an LLM for testing.
Since our method allows for many types of key generation and (stylometric) features, an exploration of further combinations, including watermark sampling strategies, will be part of the future work.

\section{Limitations and Risks}
Currently, stylometric watermarks are only implemented for the English language. In principle, our watermark can be applied to almost all languages with two restrictions based on the stylometric features we use. Acrostica require the language to have an alphabet similar in size to the Latin alphabet, so Chinese or Japanese characters would make this aspect nonfunctional.  
As for sensorimotor norms, all languages share this aspect since they relate directly to how our brains work \cite{connell_functional_2012}. A different database would be ideal for each language, although the English one by \citet{lynott_lancaster_2020} could also be translated for the same effect, since sensorimotor accuracy is not important for watermarking purposes. 

This also leads to the topic of attacks. 
While paraphrasing attacks are commonly considered in literature, other attacks are also plausible. 

Another possible limitation is as to what language models are compatible to this method. Currently, we have only shown compatibility to the popular decoder models Llama 3.1/3.2 and Mistral. However, the only architecture limitation for a language model to be compatible is to have a logits layer to manipulate generation and for the model to be sequential to be able to generate new keys.

\paragraph{Risks.}
Our work contributes to trustworthy AI, in particular by addressing the topic of accountability for proprietary language models. 
Since the key generation can be made specific for each LLM, this allows to link generated text back to its origin.
Thus, the proposed watermark and its methods have high potential to help reduce the negative risks on society by the harmful use of LLMs.

\bibliography{zotero_references,references}

\appendix
\section{Appendix}\label{sec:appendix}

\begin{table*}[ht]
  \centering
  \begin{tabular}{lccc}
    \toprule
    Configuration      & p-Senso     & p-Acro      & p-Red-Green  \\
    \midrule
    Human              & 0.60140883  & 0.99999999  & 0.94382440   \\
    Senso              & 0.00011047  & 0.99999999  & 0.88591727   \\
    Acro               & 0.63667415  & 0.00000931  & 0.94849433   \\
    Red-Green          & 0.63739815  & 0.99999999  & 0.00000102   \\
    Senso + Acro       & 0.00001441  & 0.00004255  & 0.91098964   \\
    Senso + Red-Green  & 0.00034610  & 0.99999999  & 0.00000011   \\
    Acro + Red-Green   & 0.58650698  & 0.00022102  & 0.00000004   \\
    All Three          & 0.00002030  & 0.00022102  & 0.00000001   \\
    \bottomrule
  \end{tabular}
  \label{p_values}
  \caption{Table of p-values of each feature under different ensemble settings. All values are from Llama 8B with the medium strength watermark setting.}
\end{table*}

\begin{figure*}[ht]
    \vskip 0.2in
    \begin{center}
    \centerline{\includegraphics[width=1\textwidth]{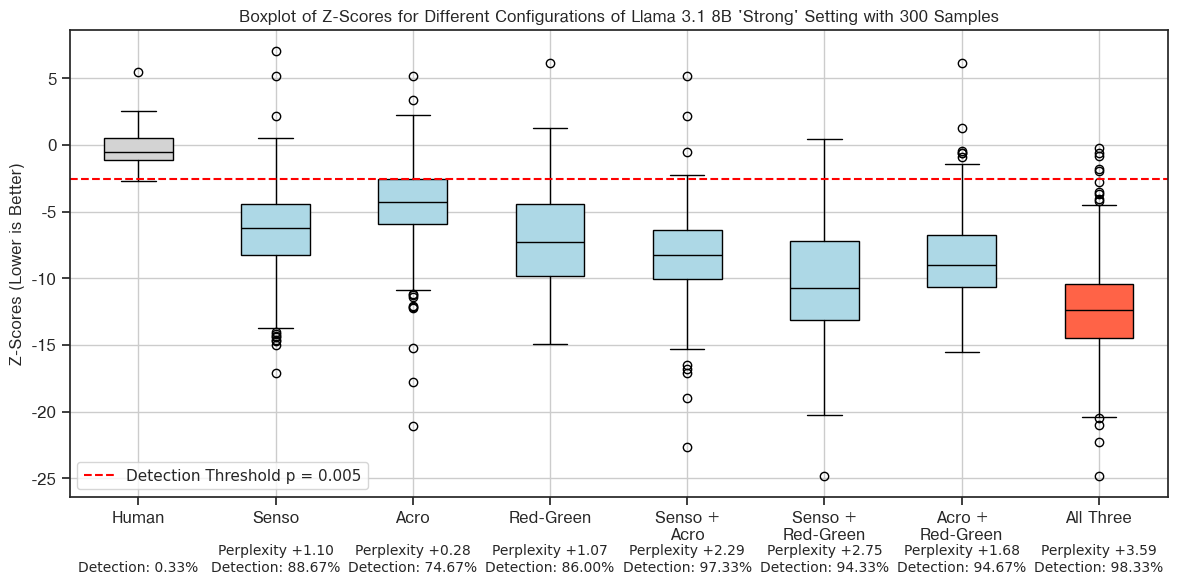}}
    \label{fig:strong}
    \end{center}
    \vskip -0.2in
\end{figure*}
\begin{figure*}[ht]
    \vskip 0.2in
    \begin{center}
    \centerline{\includegraphics[width=1\textwidth]{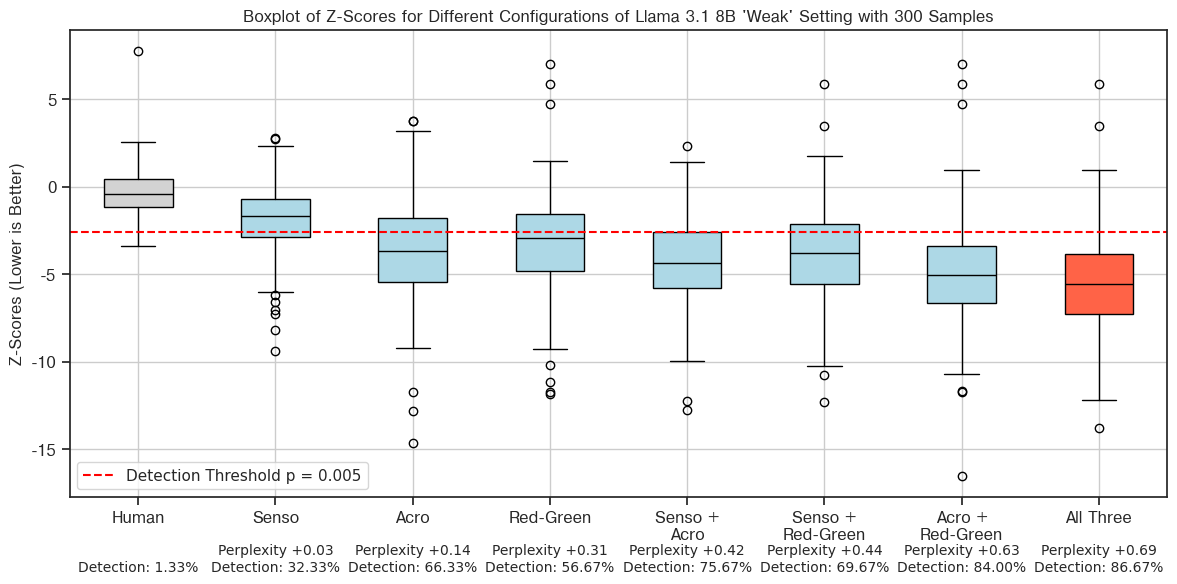}}
    \label{fig:weak}
    \end{center}
    \vskip -0.2in
\end{figure*}
\begin{figure*}[ht]
    \vskip 0.2in
    \begin{center}
    \centerline{\includegraphics[width=1\textwidth]{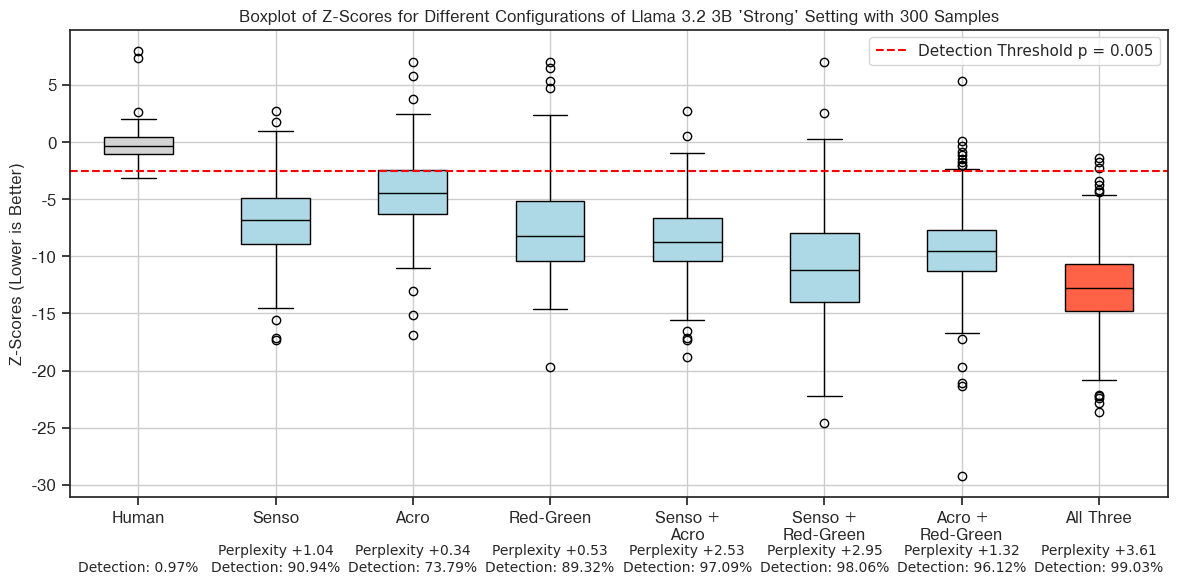}}
    \label{fig:llama3b}
    \end{center}
    \vskip -0.2in
\end{figure*}
\begin{figure*}[ht]
    \vskip 0.2in
    \begin{center}
    \centerline{\includegraphics[width=1\textwidth]{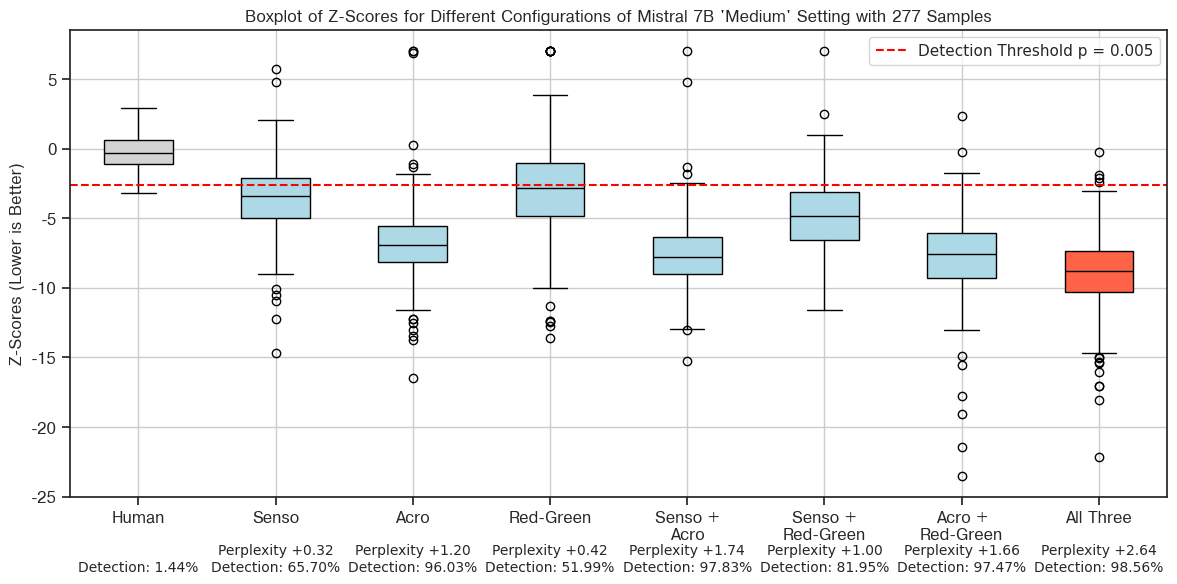}}
    \label{fig:mistral}
    \end{center}
    \vskip -0.2in
\end{figure*}
\begin{table*}[ht!]
\caption{Example of an acrostic written by Arnold Schwarzenegger in a letter to the members of the California State Assembly, which has been used in literature for illustration before \citep{stein2014generating}.}
\centering
\begin{tabular}{@{}p{\textwidth}@{}}
\toprule
To the Members of the California State Assembly: \\
\\
I am returning Assembly Bill 1176 without my signature. \\
\\
\textbf{F}\,or some time now I have lamented the fact that major issues are overlooked while many \\
\textbf{u}\,nnecessary bills come to me for consideration. Water reform, prison reform, and health \\
\textbf{c}\,are are major issues my Administration has brought to the table, but the Legislature just \\
\textbf{k}\,icks the can down the alley. \\
\\
\textbf{Y}\,et another legislative year has come and gone without the major reforms Californians \\
\textbf{o}\,verwhelmingly deserve. In light of this, and after careful consideration, I believe it is\\
\textbf{u}\,nnecessary to sign this measure at this time. \\
\\
Sincerely, \\
Arnold Schwarzenegger \\
\bottomrule
\end{tabular}
\label{tab:acrostic-schwarz}
\end{table*}

\end{document}